\documentclass[letterpaper, 10 pt, journal, twoside]{IEEEtran}
\IEEEoverridecommandlockouts

\usepackage[T1]{fontenc}

\usepackage[inline]{enumitem}
\usepackage{amsmath}
\usepackage{amssymb}
\usepackage{multirow}
\usepackage{graphicx}
\usepackage{subcaption}
\usepackage[normalem]{ulem}
\usepackage[dvipsnames]{xcolor}
\usepackage{xcolor}
\usepackage{url}

\newcommand{\AAAtodo}[1]{}
\newcommand{\AAAselfQ}[1]{}
\newcommand{\AAAfeedbackReq}[1]{}
\newcommand{\AAAnote}[1]{}

\title{
CLOVER: Context-Aware Long-Term Object Viewpoint- and Environment- Invariant Representation Learning
}
\newcommand{\modifiedAAA}[1]{#1}

\newcommand{\cutDL}[1]{}
\newcommand{\cutAAA}[1]{}
\newcommand{\lengthCutAAA}[1]{}

\author{\modifiedAAA{Amanda Adkins*} \and \modifiedAAA{Dongmyeong Lee*} \and \modifiedAAA{Joydeep Biswas}
 \thanks{\modifiedAAA{Manuscript received: March 26, 2025; Revised July 3, 2025; Accepted August 21, 2025.}}
\thanks{\modifiedAAA{This paper was recommended for publication by Editor Markus Vincze upon evaluation of the Associate Editor and Reviewers' comments. This work is partially supported by the National Science Foundation (GRFP DGE-2137420, CAREER-2046955), ARL SARA (W911NF-24-2-0025), and Amazon Lab126. Any opinions, findings, and conclusions expressed in this material are those of the authors and do not necessarily reflect the views of the sponsors.}}
\thanks{\modifiedAAA{* Equal contribution}}
\thanks{\modifiedAAA{The authors are with the Department of Computer Science, The University of Texas at Austin, Austin, TX. Email:
        {\tt\small \{aaadkins, domlee,  joydeepb\}@cs.utexas.edu }}}
 \thanks{\modifiedAAA{Digital Object Identifier (DOI): see top of this page.}}
}

\begin{document}
\maketitle
\IEEEpeerreviewmaketitle


\begin{abstract}
Mobile service robots can benefit from object-level understanding of their environments, including the ability to distinguish object instances and re-identify previously seen instances. Object re-identification is challenging across different viewpoints and in scenes with significant appearance variation arising from weather or lighting changes. Existing works on object re-identification either focus on specific classes or require foreground segmentation. Further, these methods, along with object re-identification datasets, have limited consideration of challenges such as outdoor scenes and illumination changes. To address this problem, we introduce CODa Re-ID: an in-the-wild object re-identification dataset containing 1,037,814 observations of 557 objects across 8 classes under diverse lighting conditions and viewpoints. Further, we propose CLOVER, a representation learning method for object observations that can distinguish between static object instances without requiring foreground segmentation. We also introduce MapCLOVER, a method for scalably summarizing CLOVER descriptors for use in object maps and matching new observations to summarized descriptors. Our results show that CLOVER achieves superior performance in static object re-identification under varying lighting conditions and viewpoint changes and can generalize to unseen instances and classes.

\end{abstract}



\section{Introduction}
\label{sec:intro}
\IEEEPARstart{F}{or} robotic systems to be reliable long-term, they must understand the environment at the object level, distinguishing and re-identifying object instances despite variations in viewpoints and environmental conditions. Current vision systems often struggle with these variations. For example, object-based SLAM 
systems~\cite{nicholson2018quadricslam, ok2019robust, adkins2024obvi}
use objects as landmarks; however, these methods primarily rely on geometric and semantic consistency, leading to object-level data association errors when there is trajectory drift and 
multiple instances of the same semantic class~\cite{doherty2020probabilistic, bowman2017probabilistic}.

Object re-identification is the task of recognizing an object across multiple views.
Learning-based methods address this by generating representations for images of objects and comparing a query object's representation against a database of representations for previously observed objects. 
However, most re-identification research is aimed at specific categories, such as humans~\cite{zheng2016person, ye2021deep} and vehicles~\cite{liu2016large}. Existing approaches for general object re-identification~\cite{bansal2019re, bansal2021did, kotar2024these} require foreground segmentation and focus on indoor scenes or single-object images, leaving challenges such as realistic outdoor images with significant lighting variation understudied. 

\begin{figure}
    \centering
    \includegraphics[width=0.9\columnwidth]{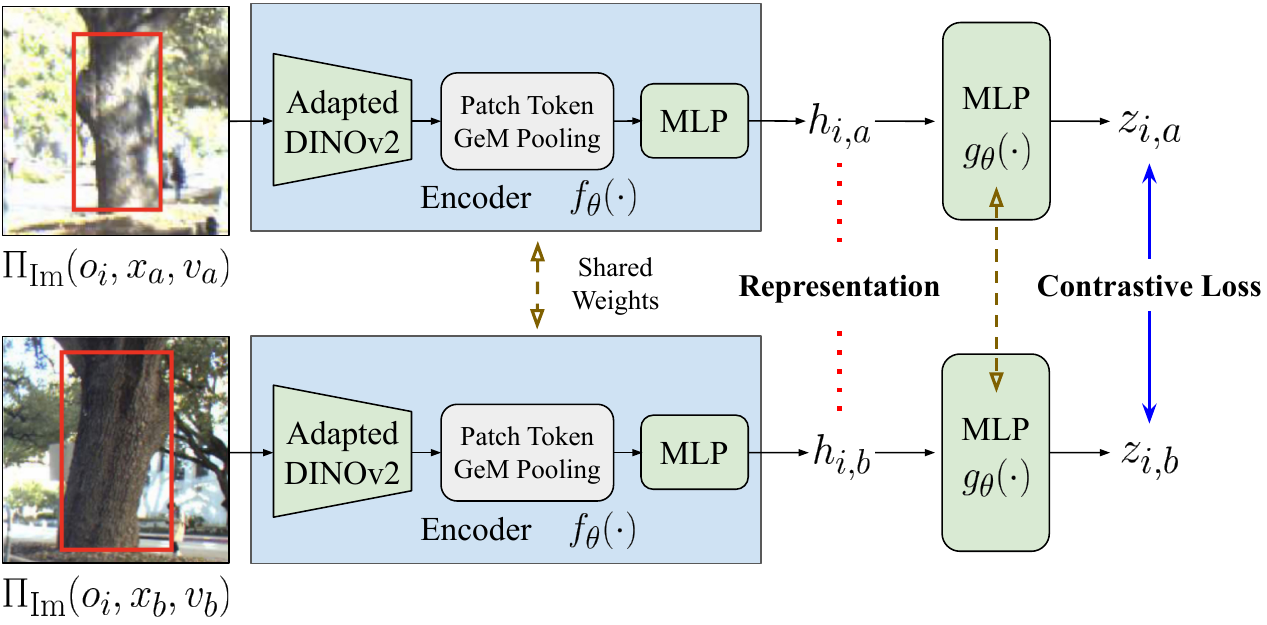}
    \caption{Architecture of  CLOVER. The tree in the input image patches is annotated with a bounding box (red) for visibility.}
    \label{fig:architecture}
    \vspace{-1.5em}
\end{figure}

In this paper, we introduce \textbf{CLOVER} (\textbf{C}ontext-aware \textbf{L}ong-term \textbf{O}bject \textbf{V}iewpoint- and \textbf{E}nvironment-Invariant \textbf{R}epresentation Learning), shown in Fig.~\ref{fig:architecture}, which extends the focus to \textit{arbitrary} static objects. CLOVER leverages a foundation model backbone and the contextual information surrounding an object as an intrinsic property of the object itself and simplifies learning by integrating object context without segmentation.
We further extend object re-identification by introducing a method to summarize object representations for use in object maps and enable comparison between summary representations and new observations. This method, called MapCLOVER, enables long-term scalability as well as improved robustness to viewpoint shifts.


Effective learning for object re-identification requires datasets capturing multiple instances from various viewpoints and environmental conditions. Existing general object datasets are limited to synthetic or indoor scenes or focus on the objects themselves. This limitation motivates the creation of the CODa Re-ID dataset, designed for general object re-identification research. Using the CODa Re-ID dataset, we evaluate existing general object re-identification methods~\cite{bansal2019re, bansal2021did, kotar2024these} and our proposed approach.

In summary, this paper makes the following contributions:
\begin{enumerate*}[label=\textbf{(\arabic*)}]
\item We introduce the CODa Re-ID dataset, containing 1,037,814 observations of 557 objects across 8 semantic classes, captured under diverse viewpoints and environmental conditions in a campus setting.
\item We introduce CLOVER, a general object re-identification architecture that leverages foundation models and contextual information of objects using supervised contrastive learning~\cite{khosla2020supervised}.
\item We introduce MapCLOVER for summarizing CLOVER descriptors for use in an object map and for matching incoming observation appearances against summary representations.
\item We compare our method with existing general object re-identification methods~\cite{bansal2019re, bansal2021did, kotar2024these} on CODa Re-ID and ScanNet~\cite{dai2017scannet} and demonstrate improved performance on both seen and unseen instances during training, varying viewpoints and lighting conditions, and unseen classes.
\end{enumerate*}


\section{Related Works}
\label{sec:related_work}
\textbf{Object Re-identification and Image Retrieval~~}
Object re-identification is the task of associating images of the same object instance captured under different viewpoints, weather, and lighting conditions. Representation learning~\cite{xiong2014person, hirzer2012relaxed}
has become standard for object re-identification.
Viewpoint- and environmental-invariant feature representation learning remains a challenge due to the visual differences of the same instance when observed from varying perspectives. To address this{ challenge}, several approaches have introduced viewpoint-aware training 
strategies~\cite{zhou2023adaptive}, 
\cite{shi2023boosting, xiahou2023identity}.
However, these works and most re-identification methods focus on specific object classes, such as person~\cite{zheng2016person, ye2021deep} and vehicle~\cite{liu2016large}. 
Few studies have extended re-identification to general objects. Foreground Feature Averaging~\cite{kotar2024these} addresses this by averaging DINOv2~\cite{oquab2023dinov2} features in the foreground and tested on varying viewpoints and lighting conditions. However, it is limited to single-object images and does not consider realistic outdoor environments. Re-OBJ~\cite{bansal2019re} and Where Did I See It (WDISI)~\cite{bansal2021did} exploit background information in addition to object appearance for re-identification. Re-OBJ jointly encodes foreground and background appearance with a CNN-based network, while WDISI uses a transformer. However, re-OBJ and WDISI only 
consider indoor scenes with limited lighting variation. Building on re-OBJ and WDISI, we investigate representation learning of 
objects by considering context as an intrinsic property of the object.

Object re-identification can be seen as a special case of image retrieval, where the goal is to retrieve similar images to a query from a database. Visual place recognition (VPR) is a related retrieval task aiming to find images from the same place as a query. SelaVPR~\cite{lu2024towards} addresses VPR by adding trainable adapters to a frozen pre-trained foundation model, enabling it to leverage 
large-scale knowledge while focusing on features most relevant to place recognition. VPR systems typically store global feature vectors of a subset of observations and may include local features for fine-grained verification or pose estimation\modifiedAAA{, though recent methods such as Placeformer~\cite{kannan2024placeformer} reduce storage by focusing on the most informative features}. There is \cutAAA{typically}\modifiedAAA{often} no aggregation of observations from
the same scene,
requiring compromises between space and retaining useful information.
Another approach for image retrieval aimed to address these challenges~\cite{wieczorek2021unreasonable} by using instance centroids instead of individual observations. MapCLOVER extends this idea, opting for a more expressive summary by clustering representations for the same object, thus finding 
a balance between storing all descriptors, naively selecting 
a subset, and oversimplifying with a single centroid.

\textbf{Deep Metric Learning~~}
Triplet loss~\cite{wang2014learning}, which minimizes the distance between representations for an anchor and positive sample and maximizes the distance from a negative sample, has been used in object
re-identification~\cite{bansal2019re, bansal2021did, bayraktar2022fast}. Self-supervised \cite{chen2020simple} 
and supervised contrastive 
learning \cite{khosla2020supervised} 
improve representation learning by leveraging multiple negatives per anchor, enabling tighter clustering and stronger separation than traditional contrastive losses like triplet loss.

\textbf{Datasets for Object Re-identification~~}
For effective representation learning and benchmarking in object re-identification, it is crucial to have datasets that capture various objects with instance labels in real environments from diverse viewpoints and environmental conditions. ScanNet~\cite{dai2017scannet}, Hyperism~\cite{roberts2021hypersim}, and Amazon Berkeley Objects (ABO) Dataset~\cite{collins2022abo} offer annotated objects but are restricted to indoor scenes, lacking real-world weather and lighting variations. The CUTE dataset~\cite{kotar2024these} includes objects captured under different lighting conditions and poses for general object re-identification tasks. However, most objects in CUTE are captured within a studio setting and lack contextual information. In contrast, our dataset includes static objects of various categories captured in outdoor scenes from multiple viewpoints, encompassing diverse weather and lighting conditions, and providing richer contextual information for real-world object re-identification.

\section{CODa Re-ID Dataset}
\begin{table*}[tb]
\vspace{0.5em}
\centering
\scriptsize
\caption{The number of object instances and observations in the CODa Re-ID dataset.}
\begin{tabular}{lcccccc}
\multicolumn{1}{c}{\multirow{2}{*}{Class}} & \multirow{2}{*}{\# instance} & \multirow{2}{*}{\# img / instance} & \multicolumn{4}{c}{\# img / instance (Scene Variation)} \\ \cline{4-7} 
\multicolumn{1}{c}{} &  &  & Sunny & Cloudy & Dark & Rainy \\ \hline
Tree & 182 & 1917.9 & 852.8 & 591.6 & 229.7 & 243.8 \\
Pole & 208 & 2141.9 & 964.1 & 667.6 & 266.2 & 244.0 \\
Bollard & 37 & 2960.1 & 1246.8 & 923.8 & 478.7 & 310.8 \\
Informational Sign & 21 & 999.1 & 435.2 & 314.1 & 115.8 & 134.0 \\
Traffic Sign & 69 & 1001.0 & 365.9 & 373.1 & 147.2 & 114.7 \\
Trash Can & 31 & 1072.0 & 467.9 & 313.1 & 152.6 & 138.4 \\
Fire Hydrant & 5 & 591.8 & 246.2 & 171.2 & 87.4 & 87.0 \\
Emergency phone & 4 & 1868.5 & 617.5 & 748.8 & 265.0 & 237.3
\end{tabular}
\label{tab:coda_stats}
\vspace{-1.5em}
\end{table*}

In this section, we describe how we augment the UT Campus Object Dataset (CODa)~\cite{zhang2024towards}  to generate the CODa Re-ID dataset, which contains globally unique \textit{object instance} labels for object observations under various lighting conditions and viewpoints. This process can be applied to other datasets, such as the NCLT dataset~\cite{carlevaris2016university} or user-generated datasets captured under varying conditions and viewpoints.

\subsection{UT Campus Object Dataset (CODa)}
UT CODa~\cite{zhang2024towards} is a robot perception dataset collected over 8.5 hours across 22 sequences. It has 1.3 million 3D bounding boxes for 53 semantic classes, along with RGB images and point clouds. The dataset includes repeated traversals of the same locations in varying weather and times of day, making CODa ideal
for generating object re-identification data under diverse lighting
conditions and viewpoints.

\subsection{Dataset Curation Process}
We leverage the geometric location of objects to generate globally unique object instance labels, assuming objects in the same location across views are the same instance.
We focus on 8 static outdoor object classes with clear spatial separability (e.g., tree, not fence) and over 500 views per instance on average; per-class details are in Table~\ref{tab:coda_stats}.
Our curation process has four steps: \begin{enumerate*}
\item global trajectory alignment, 
\item instance-level 3D bounding box annotation, 
\item bounding box projection, and 
\item segmentation and refinement.
\end{enumerate*}

\begin{figure*}[tb]
    \centering
    \begin{subfigure}[b]{0.22\textwidth}
        \centering
        \includegraphics[width=\textwidth]{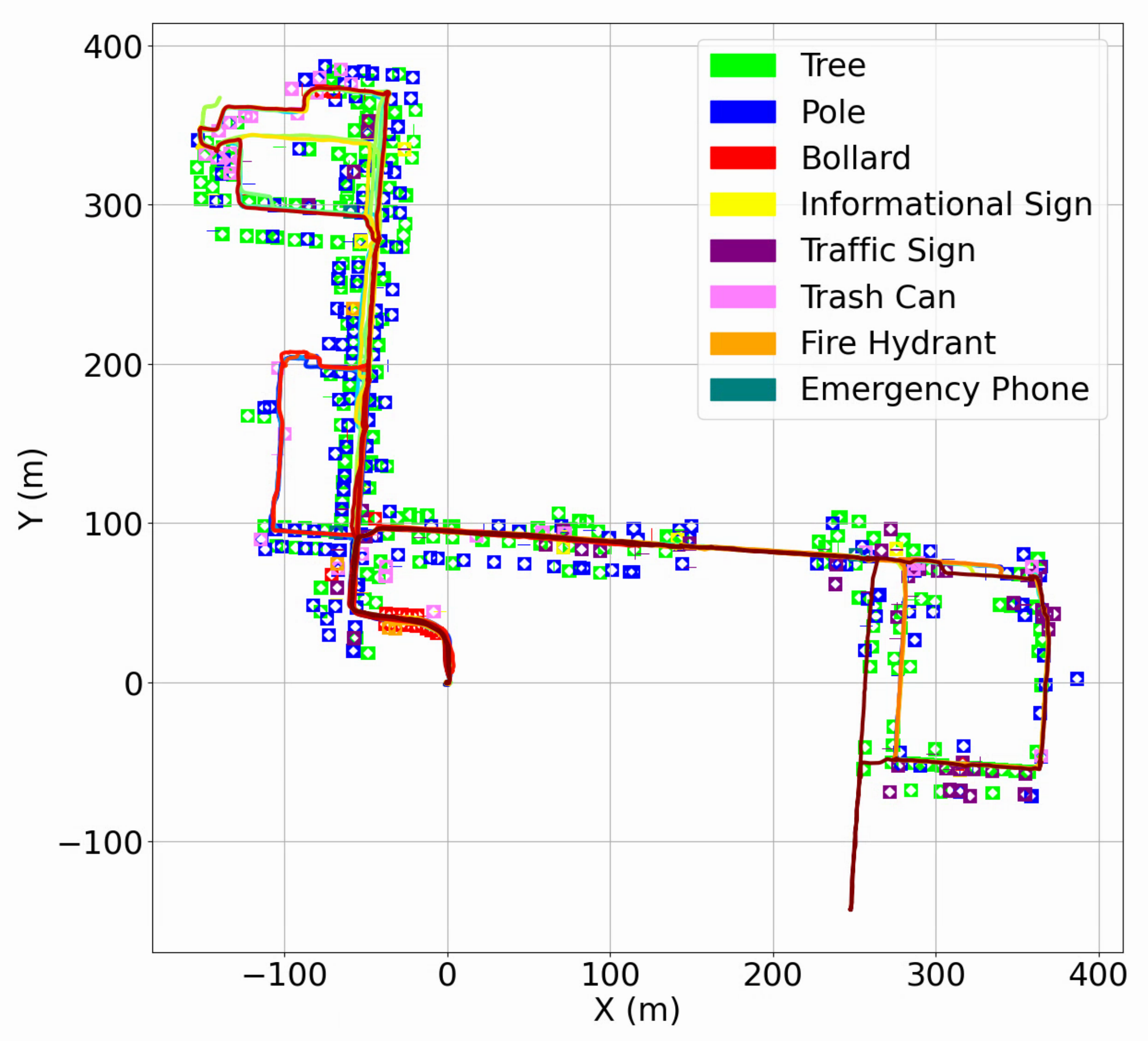}
        \caption{Globally aligned trajectories}
        \label{fig:aligned_trajectories}
    \end{subfigure}
    \hspace{1cm}
    \begin{subfigure}[b]{0.41\textwidth}
        \centering
        \includegraphics[width=\textwidth]{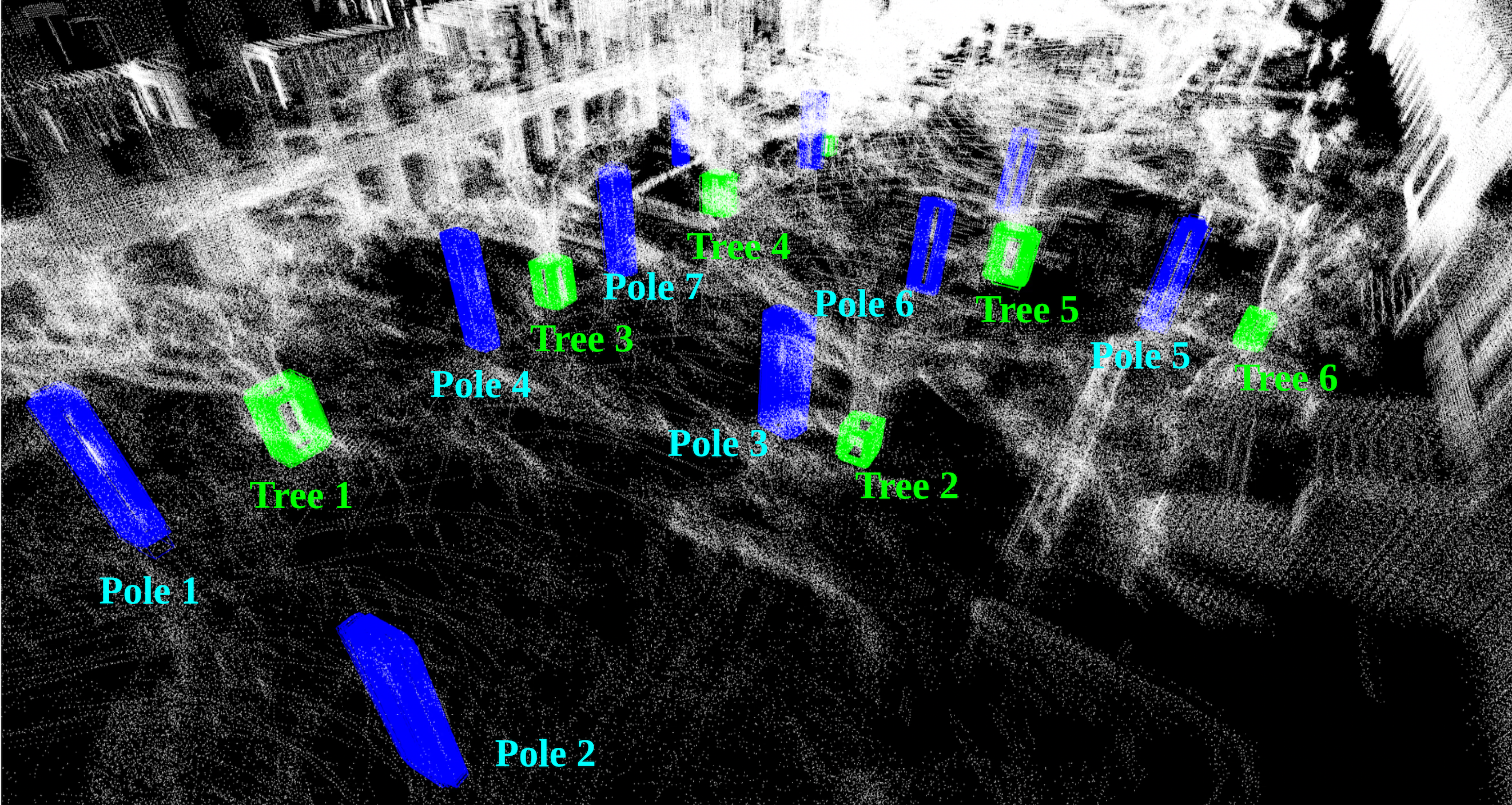}
        \caption{3D annotations for object instances}
        \label{fig:global_3d_annotations}
    \end{subfigure}
    \caption{(a) Globally aligned trajectories. (b) Global 3D bounding boxes for tree (green) and pole (blue) from CODa.}
    \label{fig:combined}
    \vspace{-1em}
\end{figure*}

\textbf{Global Trajectory Alignment~~}
As our object instance labeling relies on the location of static objects, maintaining global location consistency is necessary. After obtaining a trajectory for each sequence using LiDAR-inertial odometry~\cite{xu2022fast}, we aligned multiple trajectories of the target environment, shown in Fig.~\ref{fig:aligned_trajectories}, using interactive SLAM~\cite{koide2020interactive}, applying manual loop closures and corrections.

\textbf{Instance-level 3D Bounding Box Annotation~~} After obtaining globally consistent trajectories, we transformed the per-frame 3D bounding boxes from CODa into the global frame using the aligned sensor poses. These global-frame boxes were then clustered using DBSCAN~\cite{ester1996density} based on their 3D centroids{, considering each cluster as a single instance}. For each resulting instance, we computed a representative 3D bounding box by taking the medians of the centroids, dimensions, and rotations.\footnote{Global bounding box annotations could alternatively be annotated directly in the 3D global map.} 
The processed 3D bounding boxes in the global map are shown in Fig.~\ref{fig:global_3d_annotations}.

\textbf{Projection of 3D Annotations~~}
To generate 2D bounding box annotations with object instance labels for re-identification, we project the 3D annotations onto images. First, we query 3D annotations within a specific range around the camera pose. Next, we filter these annotations based on the camera's field of view and if point clouds are detected within the 3D bounding box at the image's timestamp. After filtering, we inscribe ellipsoids into the 3D bounding boxes and then project the ellipsoids onto the images. This reduces overestimation compared to 3D bounding boxes. For further details on ellipsoid projection, refer to \cite{nicholson2018quadricslam} and \cite{ok2019robust}.


\textbf{Instance Segmentation and Refinement~~}
Object instance foreground segmentation is required for some re-identification methods~\cite{bansal2019re, bansal2021did, kotar2024these}, so we augment CODa Re-ID with segmentation masks for greater utility. To generate masks, we used the Segment Anything Model (SAM)~\cite{kirillov2023segment}. We queried the projected 2D bounding box to obtain a segmentation mask, then adjusted the bounding box to fit the mask contour, addressing projection misalignment.

\section{CLOVER}
We introduce CLOVER, an object representation learning approach that includes context without requiring foreground segmentation. We discuss the problem formulation, image processing and loss function, and architecture for CLOVER.

\subsection{Problem Formulation}



\newcommand{\imagePatch}{\ensuremath{\mathcal{I}}}
\newcommand{\projectBB}{\ensuremath{\Pi_\mathrm{BB}}}
\newcommand{\projectIm}{\ensuremath{\Pi_\mathrm{Im}}}
\newcommand{\dataset}{\ensuremath{\mathcal{D}}}

We are interested in learning an object representation function $f_\theta$ with neural network weights $\theta$ such that the learned representations are unique to each object instance  and invariant to environmental conditions and viewpoints. Let $o_i \in O$ be the $i$th object instance, $v_a \in V$ be the $a$th environmental condition, encompassing illumination and weather, and $x_a \in X$, $X\subseteq\mathrm{SE}(3)$ be the viewpoint of the object, derived from the camera pose relative to the object. For each unique viewpoint and environmental condition, an observation of object instance $o_i$ yields an image patch $I_{i,a} \in \imagePatch$. We model this relationship with the image patch projection operator  $\projectIm : O \times V \times X \rightarrow  \mathcal{I}$, which is treated as a black-box system that produces observations \imagePatch{} for objects based on their appearance, 
environmental condition, and observing viewpoint.  The returned patch includes relevant context around the object.
$f_\theta: \imagePatch \rightarrow H$ is an object representation function that takes an observation $I_{i,a}$ and generates a representation $h_{i,a} \in H$ for that observation. Formally, we aim for $f_\theta$ to satisfy the following conditions which encode these desired uniqueness and invariance properties: 
\begin{align}
\forall v_a,v_b\in V, &x_a,x_b\in X:\qquad \label{eqn:invariance_equality} \\ &f_\theta(\projectIm(o, x_a, v_a)) = f_\theta(\projectIm(o, x_b, v_b)), \nonumber \\
 \forall o_i,o_j\in O, &i\neq j:\qquad  \label{eqn:uniqueness_equation} \\
 &f_\theta(\projectIm(o_i, \cdot, \cdot)) \neq f_\theta(\projectIm(o_j, \cdot, \cdot)). \nonumber
\end{align}

\subsection{Context-Aware Object Representation Learning}

\textbf{Contextual Object Image Patches~~} To train $f_\theta$, we need a dataset $\dataset = \left\{I_{j,a} \right\}$ consisting of observations with varying environmental conditions and viewpoints of different object instances, where $j$ is an index for instance $o_j$ and $I_{j,a}$ is an observed image patch of object $o_j$ with viewpoint and environmental condition $v_a, x_a$. We divide this dataset into {instance-specific }partitions $\dataset_i = \left\{I_{j,a}  ~|~ j=i \right\}$, each with only observations of object $i$.
As we hypothesize that limited context can help distinguish static object instances, the input $I_{i,a}$ should include context around the object. Therefore, we obtain $\{I_{i,a}\}$ by first generating object detections as 2D bounding 
boxes with center $c_{i,a}$ and dimensions $(h_{i,a}, w_{i,a})$ 
for a full image $I_\text{full}$ from a camera. For each bounding 
box, we add a fixed margin $m$ to the larger of its dimensions to get $d_{i,a} = m + \max (w_{i,a}, h_{i,a})$. Each $I_{i,a}$ is then obtained by cropping image $I_\text{full}$ into a  $d_{i,a} \times d_{i,a}$ square centered at $c_{i,a}$. This enables inclusion of context in 
the image $I_{i,a}$ for which we are generating representation $h_{i,a}$.

To improve generalization, we supplement our training procedure for $f_\theta$ by applying several augmentations~\cite{khosla2020supervised, chen2020simple, xiao2021what} to the input images $\{I_{i,a}\}$: color jittering, bounding box adjustment, crop size variation, image rotation, and random erasing. Color jittering alters color to improve lighting invariance. Bounding box position adjustment and crop size variation change the center and size of the image patch to simulate noisy bounding box detections. Image rotation simulates camera orientation variation with small random rotations, and random erasing emulates occlusions. We denote this augmented training set $\Tilde{D}$.


\textbf{Supervised Contrastive Learning~~} 
We cannot learn a function that exactly satisfies (\ref{eqn:invariance_equality}) and (\ref{eqn:uniqueness_equation}), so we instead aim to learn parameters $\theta$ such that 
\begin{align}
\forall v_a,v_b\in V, &x_a,x_b\in X, o_i,o_j \in O, i \neq j : \nonumber \\ \qquad  s(f_\theta(&\projectIm(o_i, x_a, v_a)), f_\theta(\projectIm(o_i, x_b, v_b))) \\ 
&\gg s(f_\theta(\projectIm(o_i, \cdot, \cdot)), f_\theta(\projectIm(o_j, \cdot, \cdot))), \nonumber
\end{align}
where $s(\cdot, \cdot)$ computes the representation similarity, with higher outputs 
indicating greater similarity.
To encourage similarity between representations for the same object and uniqueness across objects, we train $f_\theta$ with the augmented dataset $\Tilde{\dataset}$ using Supervised Contrastive Loss \cite{khosla2020supervised}, which enables negative information from the full batch, as opposed to triplet loss~\cite{wang2014learning}, which has only one negative sample per positive. $f_\theta$ is a neural network encoder. Following \cite{chen2020simple, khosla2020supervised, grill2020bootstrap, caron2020unsupervised}, we append a projection head $g_\theta$ to the encoder $f_\theta$ to generate the embeddings $z_{i,a}=g(h_{i,a})$ used in our loss.
This projection head $g_\theta$ maps representations to the smaller latent space where contrastive loss is applied. Our loss with temperature $\tau$ is thus given by 


\begin{equation}
    \mathcal{L} = \sum_{I_{i,a} \in \mathcal{I}} \frac{-1}{|\Tilde{\dataset}_i|} \sum_{I_{i,b} \in \Tilde{\dataset}_i, a \neq b} \hspace{-1em}\log \frac{\exp{({z_{i,a}}^\top z_{i,b} / \tau )}}{\sum\limits_{I_{j,c} \in \Tilde{\dataset}} \exp{{(z_{i,a}}^\top z_{j,c} / \tau )}}.
\end{equation} 

\subsection{Network Architecture}
We model $f_\theta$ as a neural network encoder, and define a separate projection head for contrastive learning. The encoder is derived from SelaVPR \cite{lu2024towards}, which augments a frozen DINOv2 \cite{oquab2023dinov2} backbone with learnable adapter blocks and applies GeM pooling to the generated patch tokens. To enhance encoder expressivity, we extend the adapted DINOv2 backbone with a multi-layer perceptron (MLP), which forms the final stage of $f_\theta$. This design is motivated by prior work \cite{chen2020big, caron2021emerging} showing that additional layers can improve feature quality and task performance. This dual adaptation---adapter blocks for task-specific tuning and an MLP for feature refinement---allows CLOVER to leverage DINOv2's large-scale pre-training while balancing adaptability with catastrophic forgetting risk. 
Empirically, this combination outperforms either component used in isolation. The projection head is an MLP with a single hidden layer to obtain $z_{i,a} = g(h_{i,a})$. After training, we discard the projection head $g_\theta$ and use the encoder $f_\theta$ for object re-identification. The network architecture is shown in Fig.~\ref{fig:architecture}.


\section{MapCLOVER}


\subsection{Problem Formulation}

To enable long-term scalable matching using CLOVER descriptors, we cannot simply maintain representations for all previous object observations $\{h_{i,\cdot}\}$ for each object instance $i$. Therefore, we must develop a representation summary function $f_\text{sum}$ and a summary similarity function $s_\text{sum}$.

The goal of the representation summary function $f_\text{sum}$ is to produce a summary representation $y_i$ that is bounded in size and concisely summarizes previous observations $\{h_{i,a_k}\}$ of an object instance $i$ in various viewing conditions $\{a_k\}$. This has the form $y_i = f_\text{sum}(\{h_{i, a_k}\})$.
The summary similarity function $s_\text{sum}$ should  produce a similarity score that is higher when evaluating a representation $h_{i,\cdot}$ and summary representation $y_i$ corresponding to the same object instance $i$ than when computed on a representation $h_{j, \cdot}$ and summary representation $y_i$ corresponding to different instances $i$ and $j$. Formally, the summary similarity function $s_\text{sum}(h, y)$ should satisfy the following property:
\begin{align}
    s_\text{sum}(h_{i,\cdot}, y_i) \gg s_\text{sum}(h_{j,\cdot}, y_i) \forall i, j, i \ne j. 
\end{align}
\subsection{Representation Summarization and Similarity Evaluation}

\textbf{Representation Summary Function}
The representation summary function should enable conciseness as well as tolerance of remaining variance resulting from differing viewing conditions not alleviated by CLOVER descriptor training. Averaging descriptors is not suitable in cases of notable variance, such as opposing viewpoints, as an averaged descriptor could fail to match either viewpoint well. 

To account for remaining variation in descriptors $h_{i,\cdot}$, we define the summary representation $y_i$ as a representative set of vectors $\{r_i \in H\}$ in the same space as CLOVER descriptors. Using this representative set avoids the unimodal assumptions of an averaged representation, enables intra-object variation to be captured, and prevents having to reconcile drastically different viewing conditions. Our summarization function $f_\text{sum}(\{h_{i, a_k}\})$ performs k-means clustering on all observations $\{h_{i,a_k}\}$ for an object instance $i$.  Cluster centers are used as the representative set $\{r_i\} \in y_i$.

\textbf{Summary Similarity Function~~} 
We define $s_\text{sum}$ as
\begin{align}
    s_\text{sum}(h, y_i) = \max_{r \in y_i} s(h, r),
\end{align}
where, as noted earlier, $s(\cdot, \cdot)$ computes the representation similarity. 
We use cosine similarity to align with the CLOVER training objective.  This summary similarity function computes the similarity of the query representation $h$ with each entry $r$ in the representative set and uses the highest similarity as the summary similarity.


\section{Experiments}
\label{sec:result}
We evaluate our method in real-world campus scenes with static objects. We conduct experiments on both observation-observation based matching using individual CLOVER representations and observation-map matching using summarized representations with MapCLOVER. Our CLOVER experiments address four key questions:
\begin{enumerate*}[label=(\roman*)]
    \item How invariant are representations generated by CLOVER to variations in viewpoint and lighting/weather? 
    \item How well does CLOVER generalize to unseen instances?
    \item How well does CLOVER generalize to unseen classes?
    \item What is the impact of each component of the formulation for CLOVER on its performance?
\end{enumerate*}
Additionally, we evaluate CLOVER on ScanNet \cite{dai2017scannet}. Finally, we assess how reliably incoming observations are matched to summarized representations using MapCLOVER.
CODa Re-ID, parameters, code, and extended results can be found at 
\modifiedAAA{\url{https://github.com/ut-amrl/CLOVER}}.

\subsection{CLOVER Evaluation Setup}
\textbf{Datasets~~}
To evaluate the performance of our proposed method, we trained and tested CLOVER with our CODa Re-ID dataset. The weather conditions for each object observation are defined by the sequence in which the observation was captured, categorized as sunny, dark, cloudy, or rainy. To test invariance, we created two groups of variations: illumination and viewpoint. For illumination, we generated two subsets based on the weather condition: similar illumination and different illumination. For viewpoint, for each query, we separated comparison observations into three grades of viewpoint difference difficulty based on the differences in the distance $d$ of objects from the respective camera and the angle $\alpha$ between the object-to-camera rays. We denote the subsets as easy $(d \leq 10m, \alpha \leq 15^\circ)$, medium $(d \leq 30m, \alpha \leq 90^\circ)$, and hard $(d > 30m, \alpha > 90^\circ)$. We used three classes (tree, pole, and bollard) for training across all experiments. The dataset split varied depending on the experiment, which will be explained in the following sections.

\textbf{Evaluation Metrics~~}
We quantified the performance of object re-identification using mean average precision (mAP) and top-1/top-5 accuracy, which measure the percentage of queries with a correct match in the top-1 or top-5 retrieved items. We also display the average number of matches for a query and the average retrieval set size to convey task difficulty. Object re-identification is performed among observations of objects of the same class as the query object observation, based on the assumption that an object detector can provide its class. We exclude observations of the same object from the retrieval set if they are from the same sequence as the query or if the environmental conditions or viewpoints between the query and the observations of the same object do not meet the illumination/viewpoint test case criteria.

\textbf{Implementation Details~~} 
For image patches, we use a 10-pixel margin and resize the cropped images to $224\times224$ pixels. Following SelaVPR \cite{lu2024towards}, we use the DINOv2~\cite{oquab2023dinov2} model based on VIT-L14. Our encoder MLP matches the DINOv2 output dimension of 1024 and uses one hidden layer, and our projection head MLP has an output dimension of 128 and also uses one hidden layer. We use 0.07 for our loss temperature and employ the SGD optimizer with a learning rate of 0.001. The learning rate is scheduled using CosineAnnealingLR~\cite{loshchilov2017sgdr}. We use early stopping, halting training if there is no validation mAP improvement for 10 consecutive epochs. All experiments were conducted on an Nvidia RTX A6000 GPU. CLOVER has 361 million parameters and achieves an average inference time of 15 ms per 32-image batch. In our experiments, unless otherwise noted, bounding boxes are from CODa Re-ID annotations to enable instance labeling, though CLOVER works with bounding boxes from any off-the-shelf object detector. 

\textbf{Baselines~~} 
We compare our method with re-OBJ~\cite{bansal2019re}, WDISI~\cite{bansal2021did}, and FFA~\cite{kotar2024these}. These all require segmentation, for which we used the masks provided in our CODa Re-ID dataset. We compare representations with cosine similarity for CLOVER and FFA, and $L_2$ distance for re-OBJ and WDISI, as their losses use this metric. We implemented re-OBJ and WDISI ourselves, as code was not available.

\subsection{CLOVER Experimental Results}


\begin{table*}[tb]
\vspace{0.5em}
\centering
\scriptsize
\caption{Effect of illumination and viewpoint change on object instance retrieval for object instances seen in training.}
\addtolength{\tabcolsep}{-0.25em}
\begin{tabular}{lcrr|crr|crr|crrrrrrrr}
                    & \multicolumn{3}{c|}{\multirow{2}{*}{\textbf{All}}}                                          & \multicolumn{3}{c|}{\multirow{2}{*}{\textbf{\begin{tabular}[c]{@{}c@{}}Similar\\ Illumination\end{tabular}}}} & \multicolumn{3}{c|}{\multirow{2}{*}{\textbf{\begin{tabular}[c]{@{}c@{}}Different\\ Illumination\end{tabular}}}} & \multicolumn{9}{c}{\textbf{Viewpoint Change}}                                                                                                                                                                                                                                      \\ \cline{11-19} 
                    & \multicolumn{3}{c|}{}                                                                       & \multicolumn{3}{c|}{}                                                                                         & \multicolumn{3}{c|}{}                                                                                           & \multicolumn{3}{c|}{Easy}                                                                            & \multicolumn{3}{c|}{Medium}                                                               & \multicolumn{3}{c}{Hard}                                                        \\ \cline{2-19} 
                    & mAP                                & \multicolumn{1}{c}{top-1} & \multicolumn{1}{c|}{top-5} & mAP                                      & \multicolumn{1}{c}{top-1}       & \multicolumn{1}{c|}{top-5}       & mAP                                       & \multicolumn{1}{c}{top-1}        & \multicolumn{1}{c|}{top-5}       & mAP                                & \multicolumn{1}{c}{top-1} & \multicolumn{1}{c|}{top-5}          & \multicolumn{1}{c}{mAP} & \multicolumn{1}{c}{top-1} & \multicolumn{1}{c|}{top-5}          & \multicolumn{1}{c}{mAP} & \multicolumn{1}{c}{top-1} & \multicolumn{1}{c}{top-5} \\ \hline
re-OBJ              & \multicolumn{1}{r}{0.371}          & 0.563                     & 0.712                      & \multicolumn{1}{r}{0.352}                & 0.479                           & 0.642                            & \multicolumn{1}{r}{0.314}                 & 0.445                            & 0.609                            & \multicolumn{1}{r}{0.477}          & 0.576                     & \multicolumn{1}{r|}{0.708}          & 0.204                   & 0.230                     & \multicolumn{1}{r|}{0.284}          & 0.194                   & 0.207                     & 0.419                     \\
WDISI               & \multicolumn{1}{r}{0.495}          & 0.690                     & 0.821                      & \multicolumn{1}{r}{0.480}                & 0.620                           & 0.747                            & \multicolumn{1}{r}{0.437}                 & 0.584                            & 0.730                            & \multicolumn{1}{r}{0.631}          & 0.718                     & \multicolumn{1}{r|}{0.831}          & 0.317                   & 0.356                     & \multicolumn{1}{r|}{0.533}          & 0.303                   & 0.360                     & 0.573                     \\
FFA                 & \multicolumn{1}{r}{0.236}          & 0.506                     & 0.649                      & \multicolumn{1}{r}{0.215}                & 0.391                           & 0.516                            & \multicolumn{1}{r}{0.204}                 & 0.385                            & 0.513                            & \multicolumn{1}{r}{0.393}          & 0.537                     & \multicolumn{1}{r|}{0.666}          & 0.100                   & 0.114                     & \multicolumn{1}{r|}{0.201}          & 0.083                   & 0.081                     & 0.166                     \\
CLOVER (ours)             & \multicolumn{1}{r}{\textbf{0.811}} & \textbf{0.876}            & \textbf{0.939}             & \multicolumn{1}{r}{\textbf{0.802}}       & \textbf{0.838}                  & \textbf{0.915}                   & \multicolumn{1}{r}{\textbf{0.765}}        & \textbf{0.828}                   & \textbf{0.903}                   & \multicolumn{1}{r}{\textbf{0.873}} & \textbf{0.890}            & \multicolumn{1}{r|}{\textbf{0.943}} & \textbf{0.712}          & \textbf{0.734}            & \multicolumn{1}{r|}{\textbf{0.843}} & \textbf{0.731}          & \textbf{0.777}            & \textbf{0.892}            \\ \hline
Avg Matches/Ref Set & \multicolumn{3}{c|}{15.98 / 1221}                                                           & \multicolumn{3}{c|}{8.71 / 1205}                                                                              & \multicolumn{3}{c|}{9.97 / 1213}                                                                                & \multicolumn{3}{c|}{6.27 / 1204}                                                                     & \multicolumn{3}{c|}{5.67 / 1226}                                                          & \multicolumn{3}{c}{7.62 / 1193}                                       
\end{tabular}
\label{table:seq_split_results}
\vspace{-1em}
\end{table*}

\begin{table*}[tb]
\centering
\scriptsize
\addtolength{\tabcolsep}{-0.25em}
\caption{Effect of illumination and viewpoint change on object instance retrieval for object instances not seen in training.}
\begin{tabular}{lcrr|crr|crr|crrrrrrrr}
                    & \multicolumn{3}{c|}{\multirow{2}{*}{\textbf{All}}}                                          & \multicolumn{3}{c|}{\multirow{2}{*}{\textbf{\begin{tabular}[c]{@{}c@{}}Similar\\ Illumination\end{tabular}}}} & \multicolumn{3}{c|}{\multirow{2}{*}{\textbf{\begin{tabular}[c]{@{}c@{}}Different\\ Illumination\end{tabular}}}} & \multicolumn{9}{c}{\textbf{Viewpoint Change}}                                                                                                                                                                                                                                      \\ \cline{11-19} 
                    & \multicolumn{3}{c|}{}                                                                       & \multicolumn{3}{c|}{}                                                                                         & \multicolumn{3}{c|}{}                                                                                           & \multicolumn{3}{c|}{Easy}                                                                            & \multicolumn{3}{c|}{Medium}                                                               & \multicolumn{3}{c}{Hard}                                                        \\ \cline{2-19} 
                    & mAP                                & \multicolumn{1}{c}{top-1} & \multicolumn{1}{c|}{top-5} & mAP                                      & \multicolumn{1}{c}{top-1}       & \multicolumn{1}{c|}{top-5}       & mAP                                       & \multicolumn{1}{c}{top-1}        & \multicolumn{1}{c|}{top-5}       & mAP                                & \multicolumn{1}{c}{top-1} & \multicolumn{1}{c|}{top-5}          & \multicolumn{1}{c}{mAP} & \multicolumn{1}{c}{top-1} & \multicolumn{1}{c|}{top-5}          & \multicolumn{1}{c}{mAP} & \multicolumn{1}{c}{top-1} & \multicolumn{1}{c}{top-5} \\ \hline
re-OBJ              & \multicolumn{1}{r}{0.210}          & 0.453                     & 0.617                      & \multicolumn{1}{r}{0.247}                & 0.402                           & 0.520                            & \multicolumn{1}{r}{0.183}                 & 0.381                            & 0.552                            & \multicolumn{1}{r}{0.347}          & 0.482                     & \multicolumn{1}{r|}{0.629}          & 0.070                   & 0.067                     & \multicolumn{1}{r|}{0.178}          & 0.056                   & 0.052                     & 0.127                     \\
WDISI               & \multicolumn{1}{r}{0.288}          & 0.580                     & 0.706                      & \multicolumn{1}{r}{0.364}                & 0.540                           & 0.658                            & \multicolumn{1}{r}{0.248}                 & 0.493                            & 0.635                            & \multicolumn{1}{r}{0.480}          & 0.618                     & \multicolumn{1}{r|}{0.731}          & 0.130                   & 0.143                     & \multicolumn{1}{r|}{0.255}          & 0.083                   & 0.089                     & 0.224                     \\
FFA                 & \multicolumn{1}{r}{0.294}          & 0.567                     & 0.707                      & \multicolumn{1}{r}{0.339}                & 0.492                           & 0.617                            & \multicolumn{1}{r}{0.259}                 & 0.497                            & 0.650                            & \multicolumn{1}{r}{0.446}          & 0.591                     & \multicolumn{1}{r|}{0.724}          & 0.126                   & 0.144                     & \multicolumn{1}{r|}{0.239}          & 0.121                   & 0.136                     & 0.257                     \\
CLOVER (ours)       & \multicolumn{1}{r}{\textbf{0.519}} & \textbf{0.800}            & \textbf{0.882}             & \multicolumn{1}{r}{\textbf{0.523}}       & \textbf{0.682}                  & \textbf{0.766}                   & \multicolumn{1}{r}{\textbf{0.485}}        & \textbf{0.758}                   & \textbf{0.855}                   & \multicolumn{1}{r}{\textbf{0.750}} & \textbf{0.846}            & \multicolumn{1}{r|}{\textbf{0.914}} & \textbf{0.376}          & \textbf{0.433}            & \multicolumn{1}{r|}{\textbf{0.608}} & \textbf{0.238}          & \textbf{0.261}            & \textbf{0.448}            \\ \hline
Avg Matches/Ref Set & \multicolumn{3}{c|}{13.91 / 618}                                                            & \multicolumn{3}{c|}{4.35 / 616}                                                                               & \multicolumn{3}{c|}{11.52 / 617}                                                                                & \multicolumn{3}{c|}{5.24 / 613}                                                                      & \multicolumn{3}{c|}{4.15 / 612}                                                           & \multicolumn{3}{c}{7.13 / 620}                                                 
\end{tabular}

\label{table:region_split_results}
\vspace{-1em}
\end{table*}
\begin{table*}[htb]
\centering
\scriptsize
\caption{Performance of CLOVER and comparison methods on classes not seen in training.}
\begin{tabular}{lrrrrrrrrrrr}
       & \multicolumn{3}{c}{\textbf{Informational Sign}} &  & \multicolumn{3}{c}{\textbf{Traffic Sign}} &  & \multicolumn{3}{c}{\textbf{Trash Can}} \\ \cline{2-4} \cline{6-8} \cline{10-12} 
       & mAP   & top-1   & top-5  &  & mAP    & top-1   & top-5   &  & mAP   & top-1   & top-5  \\ \cline{1-12} 

re-OBJ & 0.602 & 0.817 &   0.924 & & 0.343  & 0.607 & 0.708 &  & 0.389 & 0.678 & 0.822 \\
WDISI & 0.747 & 0.942 & 0.969 &  & 0.433 & 0.719 & 0.815 &  & 0.510 & 0.778 & 0.875 \\
FFA  & 0.735 & 0.888 & 0.929 &  & 0.391 & 0.701 & 0.765 &  & 0.505 & 0.788 & 0.859 \\
CLOVER (ours) & \textbf{0.897} & \textbf{0.964} & \textbf{0.978} &  & \textbf{0.606} & \textbf{0.840} & \textbf{0.880} &  & \textbf{0.703} & \textbf{0.850} & \textbf{0.922}  \\
\cline{1-12}
Avg Matches/Ref Set  & \multicolumn{3}{c}{22.95 / 226} &  & \multicolumn{3}{c}{23.75 / 689 } &  & \multicolumn{3}{c}{22.80 / 320}
\end{tabular}
\label{table:class_generalization}
\vspace{-2em}
\end{table*}

\textbf{Lighting, Weather, and Viewpoint Invariance~~}
To evaluate how invariant the representations are to viewpoint and lighting/weather variations, we split CODa Re-ID into a training set and a test set, with 7 and 13 sequences respectively. This is similar to a real-world deployment procedure:  initial data is collected from the target environment and used to train the model for subsequent deployments.
We narrow the retrieval set for each observation to restrict the viewpoint similarity and environmental similarity as described above to reflect robustness to the described variation.

Table \ref{table:seq_split_results} shows the results for CLOVER and the comparison algorithms when the retrieval set for each observation contains all lighting conditions, only similar illumination, only different illumination, and varying viewpoint difficulties. Cases with easy viewpoint differences have the highest performance, with performance degrading slightly as the degree of viewpoint change increases. This indicates that viewpoint shift is more challenging for all algorithms than lighting variation, but that CLOVER is still quite robust to significant viewpoint shifts. We additionally test using bounding boxes generated by YOLOv11 \cite{Jocher_Ultralytics_YOLO_2023}, obtaining instance labels from overlapping ground truth bounding boxes. The mAP of all approaches using the detections is shown in Table \ref{table:real_bb_results}; CLOVER 
again has the 
best performance.\footnote{Full results can be found on our website.}



\textbf{Unseen Instance Generalization~~}
To test CLOVER's generalization to unseen instances, we split the CODa Re-ID dataset by region such that test set instances are distinct from the training set. Table \ref{table:region_split_results} shows the results when evaluated on unseen instances in varying lighting conditions and viewpoint differences. Though the results are somewhat less accurate than for the sequence-based split in Table~\ref{table:seq_split_results}, particularly as the viewpoint difference increases, our method outperforms all baselines. This is notable as FFA~\cite{kotar2024these} is not learned, so it does not benefit from having seen instances in training.

\textbf{Unseen Class Generalization~~}
We test CLOVER's and the comparison algorithms' ability to generalize to unseen classes by evaluating on the three most common classes not used in training (informational sign, traffic sign, and trash can). 
We use weights from training using the region-based split. Table \ref{table:class_generalization} shows results for all methods. 
Despite using classes not seen in training, performance is still good, with CLOVER outperforming all methods, demonstrating generalization to unseen classes. In some cases, CLOVER and other methods have better results for unseen classes than for seen classes (Table \ref{table:seq_split_results}).
We hypothesize that this is due to a greater number of matches and more visual distinctiveness.

\begin{table}[tb]
\vspace{0.5em}
\centering
\scriptsize
\caption{Re-identification performance (mAP) on seen instances using YOLOV11 generated bounding boxes.}
\addtolength{\tabcolsep}{-0.25em}
\begin{tabular}{lccc}
       & All   & Sim. Illum.   & Diff. Illum.  \\ \cline{1-4} 
re-OBJ & 0.452 & 0.434 & 0.381 \\
WDISI & 0.585 & 0.576 & 0.516  \\
FFA  & 0.274 & 0.258 & 0.230  \\
CLOVER (ours) & \textbf{0.882} & \textbf{0.886} & \textbf{0.846} \\
\cline{1-4}
Avg Matches/Ref Set  & {14.76 / 956} & {8.32 / 942} & {9.16 / 949} 
\end{tabular}
\label{table:real_bb_results}
\vspace{-2em}
\end{table}

\begin{figure}[tb]
    \centering
    \begin{subfigure}[b]{0.21\textwidth}
        \centering
                \includegraphics[width=\textwidth]{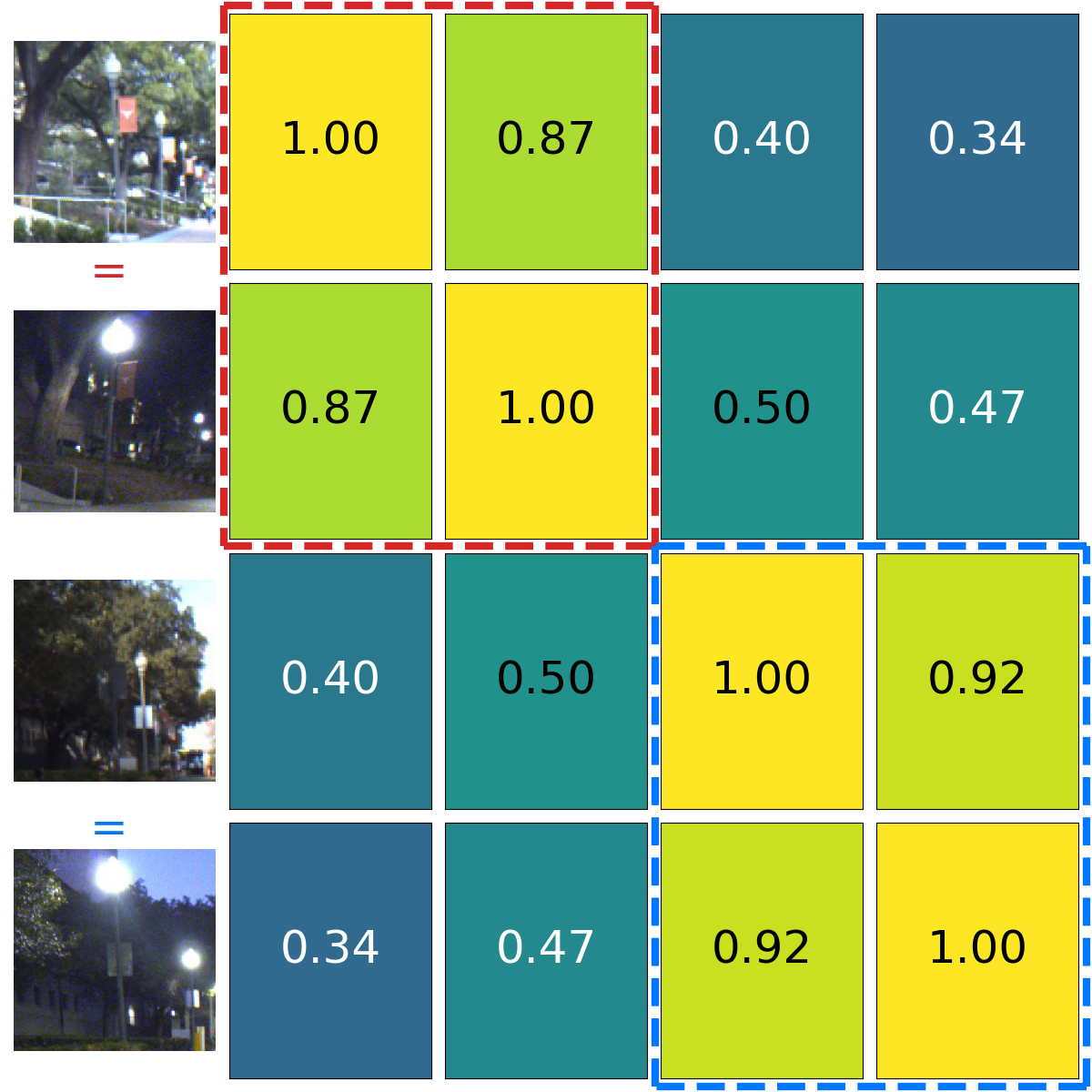}
                \subcaption{CLOVER}
        \label{fig:clover_qual}
    \end{subfigure}
    \hfill
    \begin{subfigure}[b]{0.21\textwidth}
        \centering
        \includegraphics[width=\textwidth]{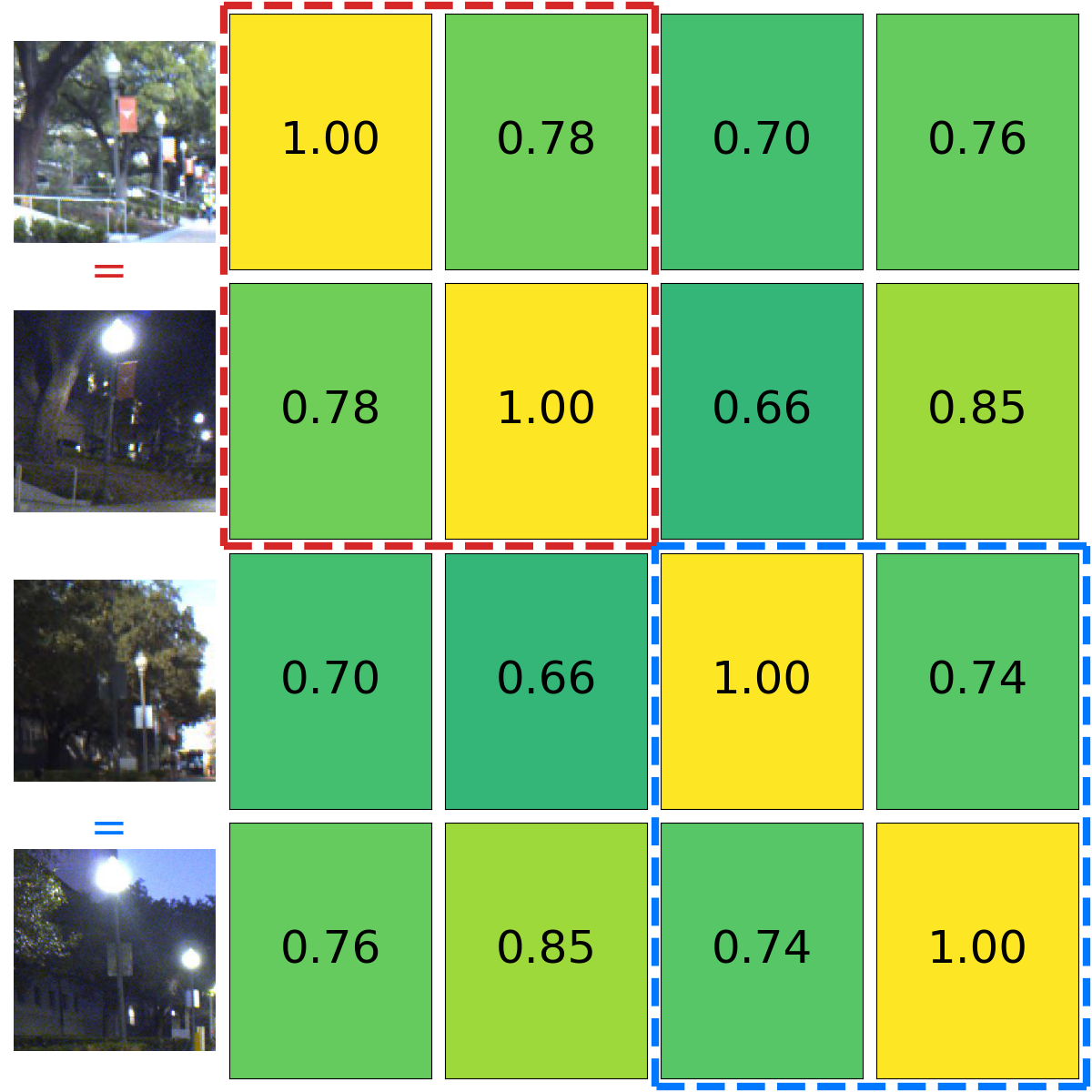}
        \subcaption{FFA}
        \label{fig:ffa_qual}
    \end{subfigure}
        \hfill
    \begin{subfigure}[b]{0.21\textwidth}
        \centering
        \includegraphics[width=\textwidth]{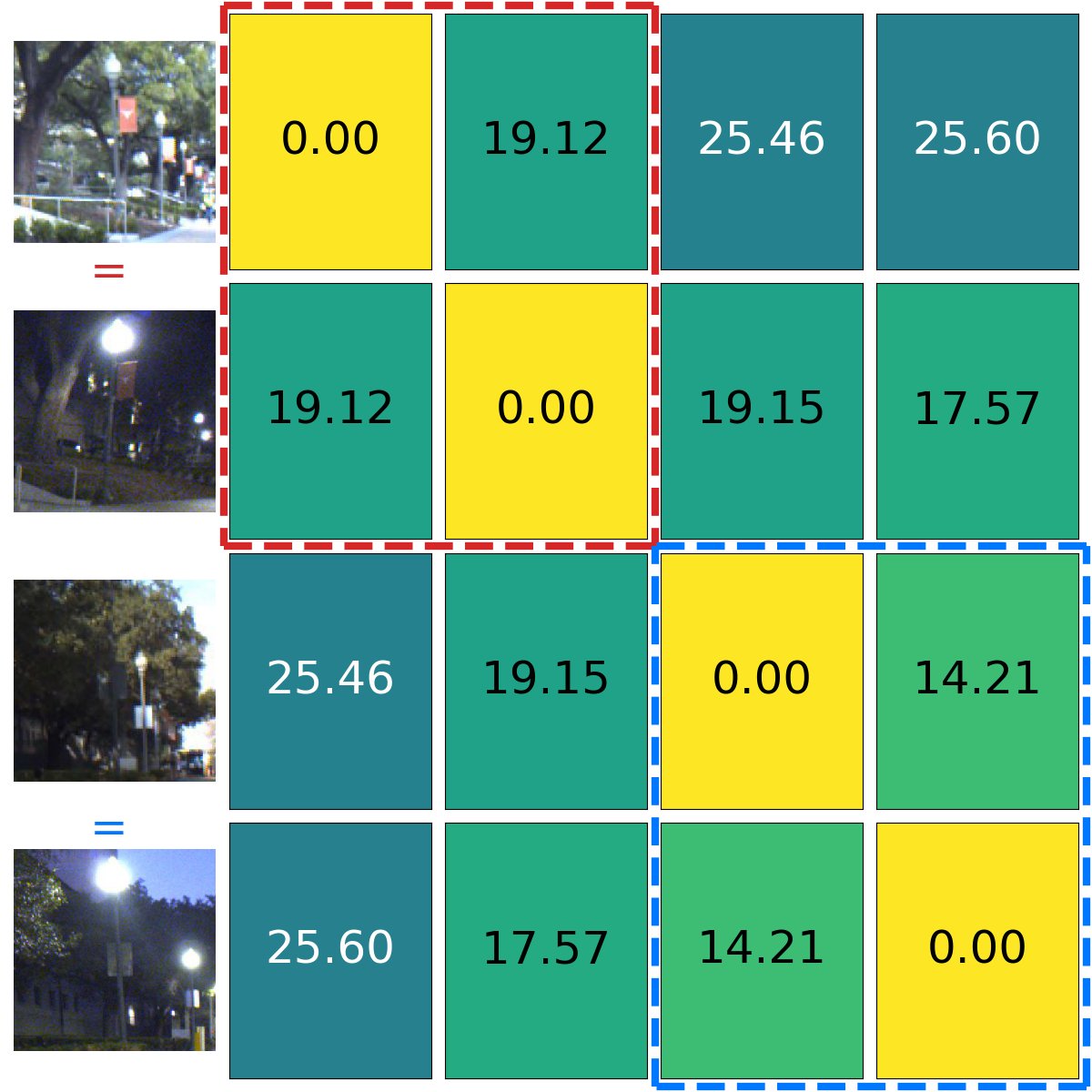}
        \subcaption{WDISI}
        \label{fig:wdisi_qual}
    \end{subfigure}
        \hfill
    \begin{subfigure}[b]{0.21\textwidth}
        \centering
        \includegraphics[width=\textwidth]{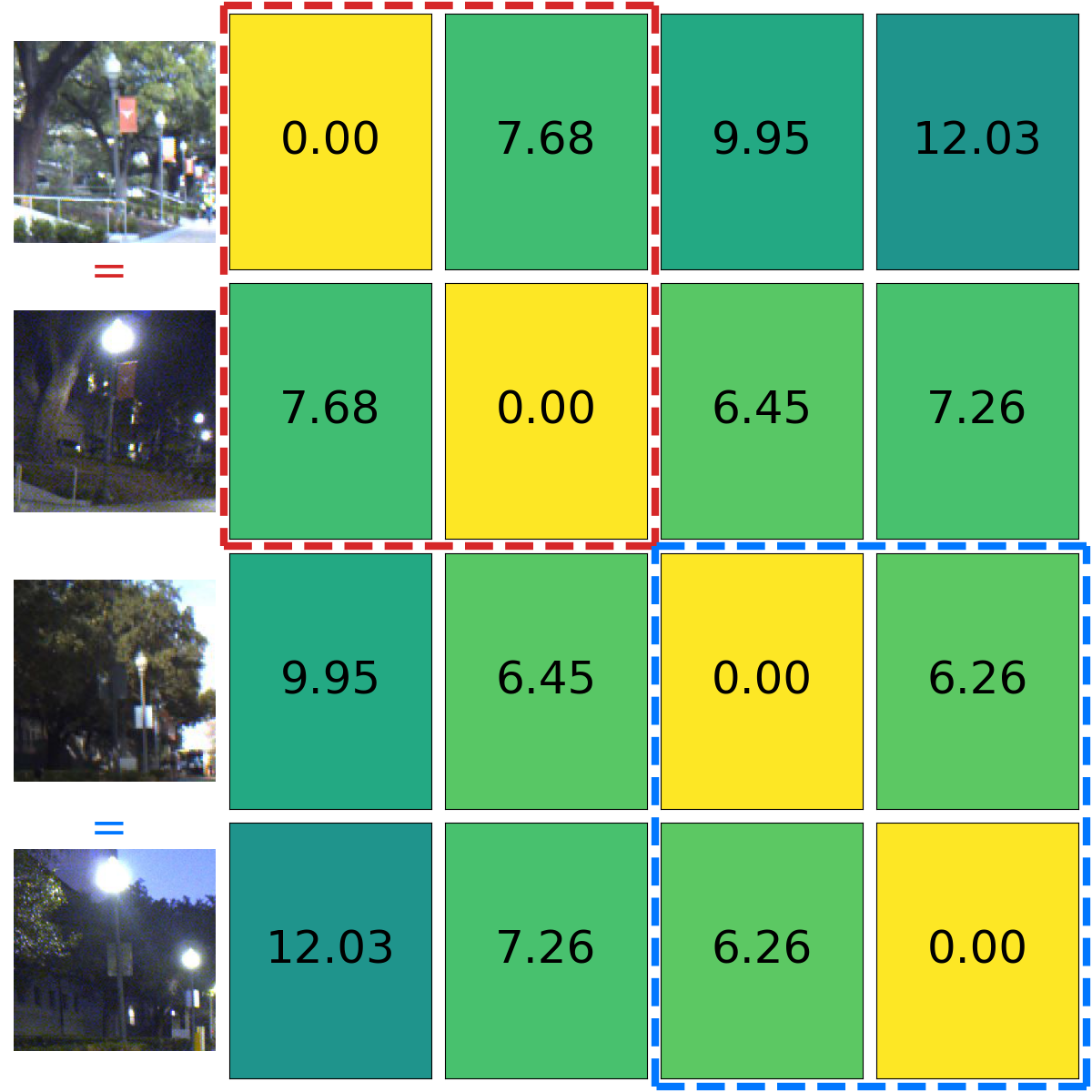}
        \subcaption{re-OBJ}
        \label{fig:reobj_qual}
    \end{subfigure}
    \caption{Qualitative performance on two pairs of images, each from the same object. Values give similarity between images: higher cosine similarity (CLOVER, FFA) and lower $L_2$ distance (WDISI, re-OBJ) indicate higher matching confidence.}
    \label{fig:qualitative}
\vspace{-2em}
\end{figure}

\begin{table*}[t]
\vspace{0.5em}
\centering
\scriptsize
\caption{Results for CLOVER and ablated versions.}
\begin{tabular}{lrrrrrrrrrrr}
       & \multicolumn{3}{c}{\textbf{All}} &  & \multicolumn{3}{c}{\textbf{Similar Illumination}}&  & \multicolumn{3}{c}{\textbf{Different Illumination}} \\ \cline{2-4} \cline{6-8} \cline{10-12} 
       & mAP      & top-1     & top-5     &   & mAP      & top-1     & top-5     &  & mAP         & top-1        & top-5        \\ \hline
CLOVER (ours) & \textbf{0.811} & \textbf{0.876} & {0.939} &  & \textbf{0.802} & \textbf{0.838} & \textbf{0.915} &  & \textbf{0.765} & \textbf{0.828} & {0.903}  \\

No Aug.      & 0.787 &  0.872 &   0.937 &       & 0.768  & 0.829 &  0.910 &  & 0.751 &  0.826 & 0.902   \\
Fixed Margin=0    & 0.766 & 0.853  & 0.920 &       & 0.757 &  0.814&  0.896& & 0.731 & 0.806 & 0.881    \\
Fixed Margin=75   & 0.803 & 0.842 & \textbf{0.947} &         & 0.783 & 0.798 & \textbf{0.915} &  & 0.756  & 0.793 & \textbf{0.922}  \\
Frozen DINOv2 (no adapters)  & 0.410 & 0.668 &  0.818 &       & 0.364 &  0.538  & 0.693 &  & 0.358 & 0.547 & 0.699 \\
Unfrozen DINOv2 & 0.496 &	0.625 &	0.737 & & 0.508 &	0.605 &	0.707	& &	0.445 &	0.539 &	0.651  \\

No Encoder MLP  & 0.765 & 0.862 &  0.928 &       & 0.746 & 0.810 & 0.894  & &  0.720 & 0.805 &  0.885 \\
Triplet Loss  & 0.563 & 0.701 &  0.832 &       & 0.541  & 0.630 &  0.771 &  & 0.498 &  0.597 & 0.738  \\
\cline{1-12}
Avg Matches/Ref Set  & \multicolumn{3}{c}{ 15.98 / 1221} &  & \multicolumn{3}{c}{ 8.71 / 1205} &  & \multicolumn{3}{c}{ 9.97 / 1213 }
\end{tabular}
\label{table:ablations}
\vspace{-2em}
\end{table*}

\textbf{Qualitative Results~~}
Fig. \ref{fig:qualitative} shows the similarity between observations of two instances in the first two and last two rows, respectively, using cosine similarity for CLOVER and FFA~\cite{kotar2024these} and $L_2$ distance for WDISI~\cite{bansal2021did} and re-OBJ~\cite{bansal2019re}. An approach correctly discriminating between object instances would have better similarity (higher for cosine, lower for distance) in the 2x2 blocks along the diagonal and worse similarity in the off-diagonal blocks. For these observations, CLOVER is the only approach that has strictly better similarity for the same instance than different instances.

\textbf{Ablations~~}
We conduct ablations with the split used in Table \ref{table:seq_split_results} to understand the impact of components and parameters on CLOVER's performance. \textit{No Aug.} tests CLOVER with no training set augmentations. We examine the impact of context by varying the margin. To understand the role of the DINOv2 backbone, we test a frozen version without adapters and an unfrozen version where the backbone weights are updated during training. We also explore architectural modifications by removing the MLP applied to aggregated DINOv2 features. The final variant uses triplet loss~\cite{wang2014learning}. Table~\ref{table:ablations} shows the performance of CLOVER and these variations, with CLOVER outperforming all others except sometimes the 75 pixel margin variant, highlighting the importance of the proposed components and parameters. The triplet loss results may also partly explain the worse performance of WDISI \cite{bansal2021did} and re-OBJ \cite{bansal2019re}, which use triplet loss. 

\begin{table}[tb]
\vspace{0.5em}
\centering
\scriptsize
\caption{Re-identification performance on ScanNet.}
\addtolength{\tabcolsep}{-0.25em}
\begin{tabular}{lrrr}
       & mAP   & top-1   & top-5  \\ \cline{1-4} 
re-OBJ & 0.466 & 0.723 & 0.853 \\
WDISI & 0.533 & 0.770 & 0.891  \\
FFA  & 0.432 & 0.662 & 0.819  \\
CLOVER (ours) & \textbf{0.658} & \textbf{0.815} & \textbf{0.931} \\
\cline{1-4}
Avg Matches/Ref Set  & \multicolumn{3}{c}{7.70 / 7273} 
\end{tabular}
\label{table:scanet_results}
\vspace{-1em}
\end{table}

\begin{table}[tb]
\centering
\scriptsize
\caption{Performance of MapCLOVER and baselines. The average number of observations per object that form the map is 8.99, and the average retrieval set size is 66.7.}

\begin{tabular}{lr|rrr}
\textbf{Summary/Sim Method}     & \multicolumn{1}{l|}{\textbf{Size}} & \multicolumn{1}{l}{\textbf{Top-1}} & \multicolumn{1}{l}{\textbf{Top-5}} & \multicolumn{1}{l}{\textbf{Top-10}} \\ \hline
Average           & -                                                & 0.764                             & 0.908                             & 0.942                              \\ \hline
Random Set, Max Sim & 5                                              & 0.738                             & 0.899                             & 0.938                              \\
Clustering, Avg Sim & 5                                               & 0.644                             & 0.856                             & 0.915                              \\
MapCLOVER (Clust, Max)          & 5                                         & \textbf{0.803}                    & \textbf{0.917}                    & \textbf{0.948}                     \\ \hline
Random Set, Max Sim & 10                                              & 0.796                             & 0.919                             & 0.949                              \\
Clustering, Avg Sim & 10                                              & 0.651                             & 0.864                             & 0.920                              \\
MapCLOVER (Clust, Max)          & 10                                      & \textbf{0.811} & \textbf{0.921}  & \textbf{0.950}    
\end{tabular}
\label{table:map_clover_results}
\vspace{-2em}
\end{table}

\textbf{ScanNet~~}To demonstrate the broader efficacy of our method, we also test on the ScanNet dataset \cite{dai2017scannet} using 240 scenes. ScanNet does not have cross-trajectory instance labels, so we train and test on disjoint subsets of the scenes, with one trajectory per included scene. All observations for each object come from the same trajectory and thus have the same lighting conditions. The test set contains 1,140 instances from 60 scenes. Table \ref{table:scanet_results} shows the mAP and top-1/top-5 accuracy of CLOVER and comparison approaches. CLOVER again outperforms all other approaches, though the improvement relative to other approaches is reduced due to the lack of challenging lighting variations. These results demonstrate that our training procedure generalizes.

\subsection{MapCLOVER Evaluation Setup}
To evaluate the reliability of MapCLOVER in matching incoming object observation appearance representations to summarized object appearance representations, we divide the test set of the region-based split used in Table \ref{table:region_split_results} into map sequences and query sequences, averaging across 30 different train/test sequence splits. We create summary descriptors for each object instance using CLOVER descriptors for observations in the map sequences and use descriptors for query sequence observations as queries. We compare MapCLOVER against two baselines: \begin{enumerate*}[label=(\arabic*)]
\item averaging all map sequence CLOVER descriptors to get a summarized representation, similar to \cite{wieczorek2021unreasonable}, and
\item randomly selecting descriptors to form a representative set. \end{enumerate*}  We also evaluate using the average similarity as our summary similarity function, instead of the maximum. Additionally, we test representative set sizes 5 and 10 to understand the tradeoff between accuracy and scalability. Performance is assessed using top-1, top-5, and top-10 accuracy.

\subsection{MapCLOVER Experimental Results}

Table \ref{table:map_clover_results} shows the top-k metrics for random representative set, observation averaging, and MapCLOVER, using both average and maximum similarity for comparison. MapCLOVER has notably improved accuracy compared to randomly choosing a representative set  or averaging. Further, using the maximum similarity as proposed yields better accuracy than average similarity. Additionally, accuracy improves as the representative set size increases, but MapCLOVER's advantage over other approaches is greatest at smaller set sizes, demonstrating MapCLOVER's ability to enable concise representations with minimal accuracy impact.


\section{Conclusion}
\label{sec:conclusion}
In this paper, we propose CLOVER to generate representations for object observations, enabling discrimination between object instances with improved environment condition and viewpoint invariance compared to existing methods. MapCLOVER builds upon CLOVER by providing methods to concisely augment an object map with summary appearance  data and match object summaries with incoming observations. Our CODa Re-ID dataset includes observations in varying lighting conditions and viewpoints to enable further research into general object re-identification. 

CLOVER, which leverages limited context to re-identify static objects, is not suitable for dynamic objects like humans and vehicles, as their backgrounds can change unpredictably. 
Future work could include utilizing CLOVER-generated representations to improve object data association in object-based SLAM for both short- and long-term changes.
Finally, a joint matching framework to consider compatibility of both geometry and appearance representations for multiple object observations simultaneously could enable more robust association using CLOVER and MapCLOVER.




\bibliographystyle{IEEEtran}
\bibliography{references}  

\clearpage

\end{document}